\documentclass[]{spie}  %>>> use for US letter paper
\usepackage{amsmath,amsfonts,amssymb}
\usepackage{graphicx}
\usepackage{subfig}
\usepackage[export]{adjustbox}
\usepackage[colorlinks=true, allcolors=blue]{hyperref}
\usepackage{booktabs}
\usepackage[style=base]{caption}
\usepackage{floatrow}
\newfloatcommand{capbtabbox}{table}[][\FBwidth]

\usepackage{varwidth}
\DeclareCaptionFormat{myformat}{%
  \begin{varwidth}{\linewidth}%
    \centering
    #1#2#3%
  \end{varwidth}%
}
\title{Multi-Objective Dual Simplex-Mesh Based Deformable Image Registration for 3D Medical Images -- Proof of Concept}

\author[a]{Georgios Andreadis}
\author[b]{Peter A. N. Bosman}
\author[a]{Tanja Alderliesten}
\affil[a]{Dept. of Radiation Oncology, Leiden University Medical Center (LUMC), P.O. Box 9600, 2300 RC Leiden, The Netherlands}
\affil[b]{Life Sciences and Health Group, Centrum Wiskunde \& Informatica (CWI), P.O. Box 94079, 1090 GB Amsterdam, The Netherlands}

\authorinfo{Address all correspondence to: Georgios Andreadis. E-mail: G.Andreadis@lumc.nl}

\begin{document}
\maketitle

% 250 word limit
\begin{abstract}
Reliably and physically accurately transferring information between images through deformable image registration with large anatomical differences is an open challenge in medical image analysis.
Most existing methods have two key shortcomings: first, they require extensive up-front parameter tuning to each specific registration problem, and second, they have difficulty capturing large deformations and content mismatches between images.
There have however been developments that have laid the foundation for potential solutions to both shortcomings.
Towards the first shortcoming, a multi-objective optimization approach using the Real-Valued Gene-pool Optimal Mixing Evolutionary Algorithm (RV-GOMEA) has been shown to be capable of producing a diverse set of registrations for 2D images in one run of the algorithm, representing different trade-offs between conflicting objectives in the registration problem.
This allows the user to select a registration afterwards and removes the need for up-front tuning.
Towards the second shortcoming, a dual-dynamic grid transformation model has proven effective at capturing large differences in 2D images.
These two developments have recently been accelerated through GPU parallelization, delivering large speed-ups.
Based on this accelerated version, it is now possible to extend the approach to 3D images.
Concordantly, this work introduces the first method for multi-objective 3D deformable image registration, using a 3D dual-dynamic grid transformation model based on simplex meshes while still supporting the incorporation of annotated guidance information and multi-resolution schemes. 
Our proof-of-concept prototype shows promising results on synthetic and clinical 3D registration problems, forming the foundation for a new, insightful method that can include bio-mechanical properties in the registration.
\end{abstract}


\section{INTRODUCTION}
\label{sec:introduction}

Many clinical treatments require information to be transferred between multiple medical images of the same patient.
These images can feature large deformations and content mismatches, which call for a deformable image registration method to derive a likely spatial correspondence.
Existing methods typically require the user to fine-tune parameters (e.g., the weights of different objectives in an optimization function) before returning a single registration solution to the user.
The problem of finding appropriate weights of objectives for specific patients and applications is, however, difficult and not generally solvable~\cite{Pirpinia2017}.
Existing methods also struggle to capture large deformations and content mismatches.
Current applications therefore need to rely on tuning parameter settings for deformable registration methods by trial and error or need to fall back on more simplistic rigid registration.

The problem of up-front parameter tuning can be addressed by taking a multi-objective approach, removing the need for such tuning by exploring the problem space and producing a set of high-quality solutions that represent the trade-off between conflicting objectives.
These solutions are non-dominated, i.e., no solution exists in this set with at least a better value for one of the objectives while having similar values for the other objectives.
From this so-called approximation set, the user can select a solution that fits the current application.
This multi-objective strategy therefore incorporates user domain knowledge into the registration process.

The problem of large deformations and content mismatches has recently been addressed for 2D images through a dual-dynamic grid transformation model~\cite{Alderliesten2013}.
Contrary to the conventional transformation model of a fixed grid overlaid on one image and a dynamic (moving) grid on the other image, this approach employs two dynamic grids to flexibly match structures in both images.
Prior work has shown that Evolutionary Algorithms (EAs) can be effective at solving the associated multi-objective optimization problem for 2D registration~\cite{Alderliesten2015}.
By exploiting local dependencies in the problem, the optimization process can be sped up significantly through the partial evaluation mechanism in the Real-Valued Gene-pool Optimal Mixing Evolutionary Algorithm (RV-GOMEA)~\cite{Bouter2017}.
This has recently been accelerated even further by adapting the algorithm for a Graphics Processing Unit (GPU)~\cite{Bouter2021},
thereby opening the door to the extension of this type of deformable image registration to 3D images.

While previous work has shown promising results for multi-objective deformable registration in 2D, no multi-objective method exists yet for 3D images.
The third dimension presents a number of new challenges, such as the drastically increased parameter count due to the added dimension and the increased complexity of grid element operations (e.g., voxel iteration, collision detection).
In this work, we present the first extension for 3D images.
We present a 3D model for dual-dynamic grid transformations that guarantees inverse-consistent registration.
Our model also supports multi-resolution registration, to accommodate both large and small content changes.
We study the performance of our proof-of-concept prototype on a  synthetic and a clinical example featuring large deformations.

\section{MATERIALS AND METHODS}
\label{sec:materials-and-methods}

\subsection{Dual-dynamic grid transformation model}
A registration of a source image onto a target image requires finding a likely spatial correspondence between the two images.
A common approach to this problem is to overlay the source and target images with regular grids and to deform the grid on the target image to model the transformation~\cite{Crum2004}.
The contents of one grid cell on the source image can then be mapped to the contents of the corresponding grid cell on the target image.
In the dual-dynamic grid transformation model~\cite{Alderliesten2013}, 
the grid on the source image can also be deformed, allowing the model to naturally and efficiently capture large deformations and content mismatches.
For example, a disappearing structure can be modeled with large grid cells on the structure in the source image and small corresponding grid cells in the target image where the structure has disappeared.
However, this flexibility comes at the cost of increased complexity, as it requires twice as many parameters to be optimized.

\begin{figure}
    \centering
    \includegraphics[width=\textwidth]{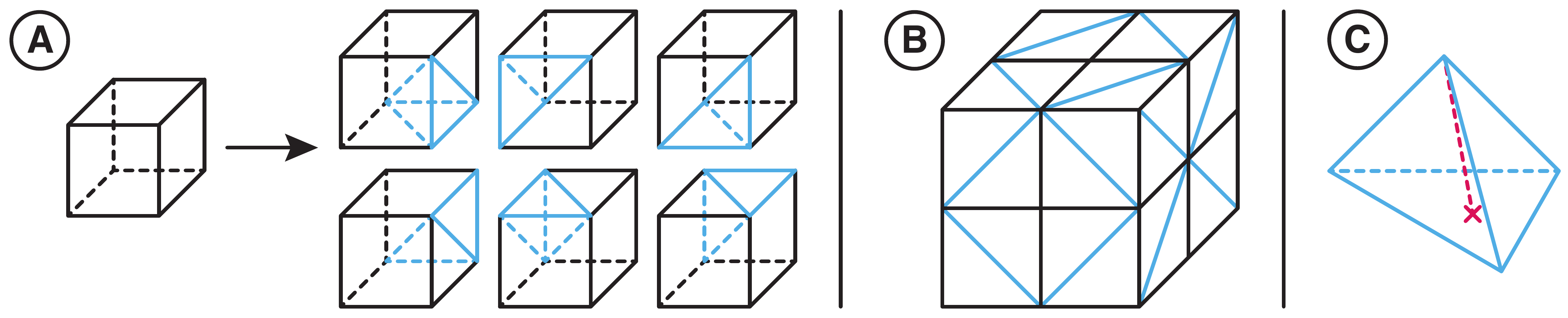}
    \caption{A grid transformation model for free-form deformations. (A) Division of a cube grid cell formed by 8 neighbour points in the grid mesh into 6 tetrahedra (in blue). (B) Symmetric alignment of groups of 8 cubes to reduce transformation bias. (C) Addition of a spoke edge (in red) inside of a tetrahedron to capture collapsing motion.}%\vspace{-0.4cm}}
    \label{fig:tetrahedral-division}
    %\vspace{-0.2cm}
\end{figure}

In previous work that uses this transformation model on 2D images~\cite{Bouter2017}, the grid points are connected to form regular triangulated topologies.
This divides the space into simplices -- shapes with the minimum number of vertices to form an $n$-dimensional volume.
In this work, we translate the transformation model to 3D image space, in which the tetrahedron is the simplex shape.
Inspired by a component of the marching tetrahedra rendering algorithm~\cite{Doi1991},
we form 3D grid meshes and subdivide them by forming 6 tetrahedra per set of 8 grid points, as illustrated in Figure~\ref{fig:tetrahedral-division} (A).
By dividing all cube grid cells with this strategy, we form a regular grid topology.
This topology is not yet axis-symmetrical, however, which can lead to unrealistic transformation biases towards certain directions.
To address this, we consider groups of 8 adjacent cube grid cells and within each group align the cells in a mirrored fashion, as shown in Figure~\ref{fig:tetrahedral-division} (B).

% Why EA: MO & flexibility to use non-smooth models and objectives

\subsection{Multi-objective optimization on the GPU with RV-GOMEA}
This work builds on prior work on multi-objective 2D registration with RV-GOMEA~\cite{Bouter2021}.
Unlike classic EAs, the RV-GOMEA algorithm can exploit the independence of non-neighbouring image regions to explore several changes to subsets of variables in parallel, on the GPU.
Each such partial change is only accepted if it improves the quality of the solution.
Per cube grid cell, one of its six tetrahedra is selected as a set of points that are modified together in a partial change.
The problem dependency structure is made explicit by identifying the tetrahedra that need to be re-evaluated when the coordinates of a certain grid point change.

\subsection{Optimization objectives}
We consider three objectives of interest (to be minimized): 1) the dissimilarity in intensity values; 2) amount of deformation; and 3) guidance information.
The last objective is optional and can be left out when no guidance information is available.
Each objective is decomposable as a sum of values made up of partial values for each grid cell, making efficient evaluations of partial solution changes possible.
We also subject the optimization to hard constraints to limit the search to feasible registrations (i.e., we prevent grid folding).

\subsubsection{Dissimilarity objective}
The dissimilarity objective captures how well the transformed image matches the actual image.
While it is well-known that measures like mutual information and cross correlation have valuable properties for deformable image registration, for simplicity of our proof-of-concept, we use a measure that is straightforwardly compatible with partial evaluations.
Note, this does not mean that using the other measures is not compatible with our approach.
We use the mean squared difference of intensity values per pair of corresponding image voxels.
In the case a black (empty) voxel is matched with a grey (non-empty) voxel, the maximal grey-value difference of 1 is used to encourage the matching of volume contours.
The mean squared difference is computed both in the forward direction (transformed source compared to target) and backward direction (transformed target compared to source) to account for potential content mismatches.
The total sum is normalized by the total number of voxels so that the metric is independent of the image resolution.
We rasterize each tetrahedron into voxels by computing the intersections of axial slices with the tetrahedron and then iterating over the voxels that lie inside the 1 or 2 intersection triangles in a scanline fashion.
The edge case of voxels lying on edges between tetrahedral slices and being associated to both tetrahedra is addressed using the top-left rasterization rule\footnote{The top-left rule in rasterization states that a pixel lies inside a triangle if positioned on the top edge (exactly horizontal and above other edges) or left edge (not horizontal and on the left side) of that triangle.}.

\subsubsection{Deformation magnitude objective}
The deformation magnitude objective represents the amount of energy needed to perform the transformation from the source grid to the target grid.
This measure is based on Hooke's law using the sum of differences in edge lengths between source and target grid edges~\cite{Arfken1985}.
To also capture changes resulting in flat tetrahedra, where one vertex has moved close to the opposite side formed by the other three vertices, we add spoke edges between each vertex and the centroid of its opposing side (see Figure~\ref{fig:tetrahedral-division} (C)).
We normalize by the total number of edges.
% \todome{Add formula when it is clear which exact energy works and if enough space}

\subsubsection{Guidance error objective}
The guidance error objective allows the user to supply additional structural knowledge in the form of corresponding contours or landmarks.
Each correspondence is represented by a set of points on the source image and a set of points on the target image.
For each point in a guidance set, the smallest distance to any point after transformation is computed (in a symmetric fashion and normalized by dividing by the total number of points).
The sum over all the squared distances is to be minimized\cite{Alderliesten2015}.

\subsubsection{Optimization constraints}
To prevent folding of grids, we check whether any grid points are located inside any grid cells that they do not belong to.
Generally, grid points are checked for collision with tetrahedra that they do not belong to as follows: each point is surrounded by a 3D polygon mesh of triangular faces of surrounding tetrahedra.
A constraint violation is defined as the point being outside of this polygon, checked through a series of ray-to-triangle intersections using a fast intersection algorithm~\cite{Jimenez2010}.
An exception to this strategy holds for grid points that are located on an image border.
These are constrained to only move across this 2D border and are thus checked for collision as follows:
each border point is surrounded by a 2D polygon of line segments belonging to surrounding tetrahedra that the point does not belong to.
A constraint violation is again defined as the point being outside of the polygon and checked by a series of ray-to-line-segment intersections.

\subsection{Grid multi-resolution schemes}
To counteract convergence to local optima, we use a hierarchical, multi-resolution scheme where registration solutions obtained at a coarser resolution are used as starting point for registration at a finer resolution~\cite{Bajcsy1989,Lester1999}.
We refine existing solutions by adding points along the existing grid meshes associated with the solutions, thereby preserving the feasibility of solutions.
For each cube grid cell (consisting of 6 tetrahedra) in both the source and the target grid, we place points in the middle of each edge and in the middle of each face, as well as in the center-of-mass of the cube.
Each cube of 8 points is thus subdivided into 8 cubes of 27 points in total, analogous to an existing approach for 2D images\cite{Alderliesten2014}. 
A finer resolution run is initialized with a random sample of the solutions taken from the approximation set (which is generated from an archive of non-dominated solutions encountered during optimization) of the previous, coarser resolution run.
The size of this random sample is equal to the population size of the new run: each individual in the new population is derived from an individual of the run at a coarser solution.

\subsection{Experiments}
To examine whether our proposed approach is able to successfully register 3D images with large deformations, we consider two registration problems.
First, we create a synthetic registration problem that simulates a cube being deformed in a non-affine fashion, with a shrinking sphere in the center.
Each side of the cube is deformed in the shape of a 3D parabola orthogonally intruding into the cube towards the center.
We provide guidance information consisting of points on the shrinking sphere surface.
Second, we register two images of a strongly deformed bladder, as a proof-of-concept test for clinical registration. 
For this, we retrospectively select two CT scans of the abdominal area of a cervical cancer patient acquired for radiation treatment planning purposes.
On the first scan, the bladder is filled, and on the second scan, taken shortly after, the bladder is empty and thus has shrunken significantly.
A radiation therapy technologist manually annotated organ segmentations on both scans, for use as guidance information.
These clinical CT scans are pre-processed before deformable registration.
First, we resample the images to a voxel spacing of $(3, 3, 3)$ (in $mm$) for consistent resolutions, matching the thickness of the axial slices of the CT scans ($3mm$). 
Then, we apply affine registration with the Elastix toolkit for initial alignment of the bony anatomy.
Following this alignment, we segment out the bladder per axial slice, using a 2D polygonal estimate based on the manual annotations provided by the radiation therapy technologist.
Finally, we crop the images to fit the largest bladder including a margin.

Both experiments rely on the input of guidance information to converge to high-quality and relevant solutions.
We assume these annotations to be present even in live clinical settings, where they do not need to be manually created but could be generated with automatic organ contouring software.
This is in line with related work, which also often assumes organ contours to be given as input to the registration process.~\cite{Zhong2010,Kaus2007}
For both experiments, we use a multi-resolution scheme with grids of size $6 \times 6 \times 6$ and $11 \times 11 \times 11$, corresponding to $5 \times 5 \times 5$ and $10 \times 10 \times 10$ cubes in each grid mesh, respectively.
This results in 1296 and 7986 parameters to be optimized, respectively, as the number of grid points (e.g., $6*6*6$) is multiplied by the number of coordinates per point (3) and the number of grids (2, as this is a dual-dynamic transformation model).
Our experiments are executed on a Dell PowerEdge R740 node with two Intel Xeon Platinum 8160 CPU cores and one NVIDIA TITAN Xp GPU (3840 CUDA cores).
Execution terminates after 300 generations for the first resolution and 600 generations for the second, with 250 individuals in the first population and 500 in the second population.
These settings have been empirically determined to allow for convergence to a diverse set of high-quality solutions.

\newcommand{\imageHeight}{3.9cm}
\newcommand{\dvfHeight}{4.4cm}

\begin{figure}
    \centering
    \subfloat[\centering Problem:\newline Source image]{\includegraphics[height=\imageHeight,valign=c]{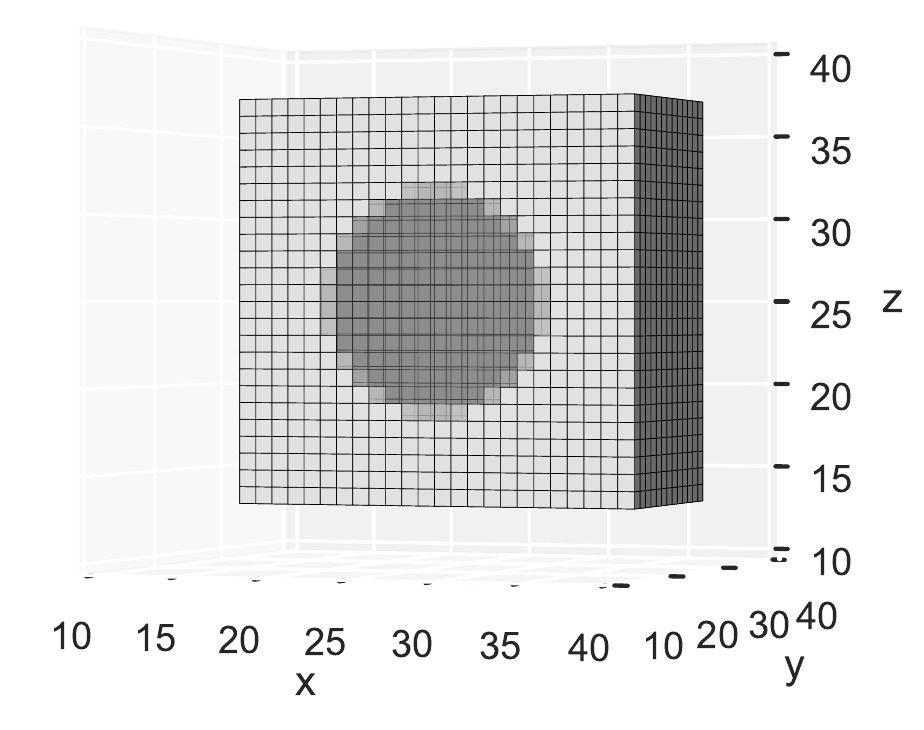}}\quad
    \subfloat[\centering Problem:\newline Target image]{\includegraphics[height=\imageHeight,valign=c]{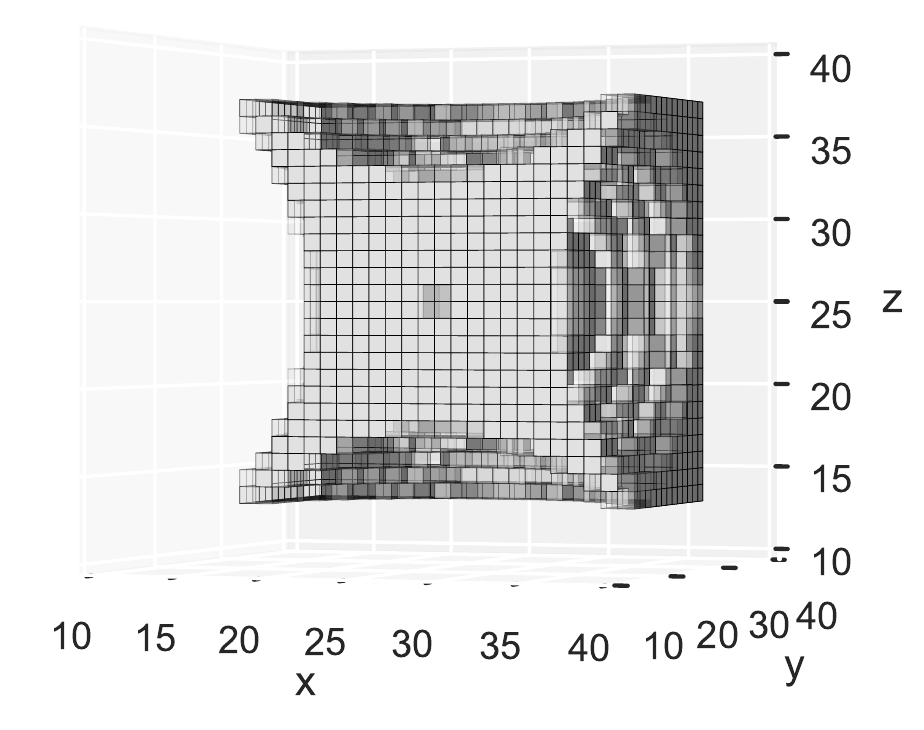}}\quad
    \subfloat[\centering Solution:\newline Transformed source]{\includegraphics[height=\imageHeight,valign=c]{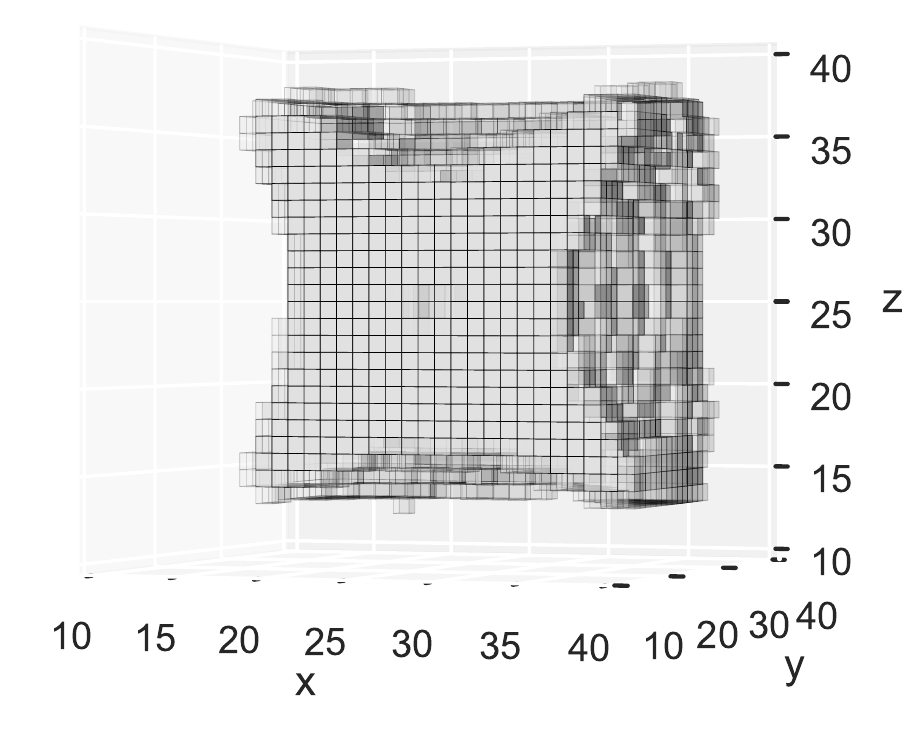}}\quad
    \subfloat[\centering Solution:\newline Axial DVF Slice]{\includegraphics[height=\dvfHeight,valign=c]{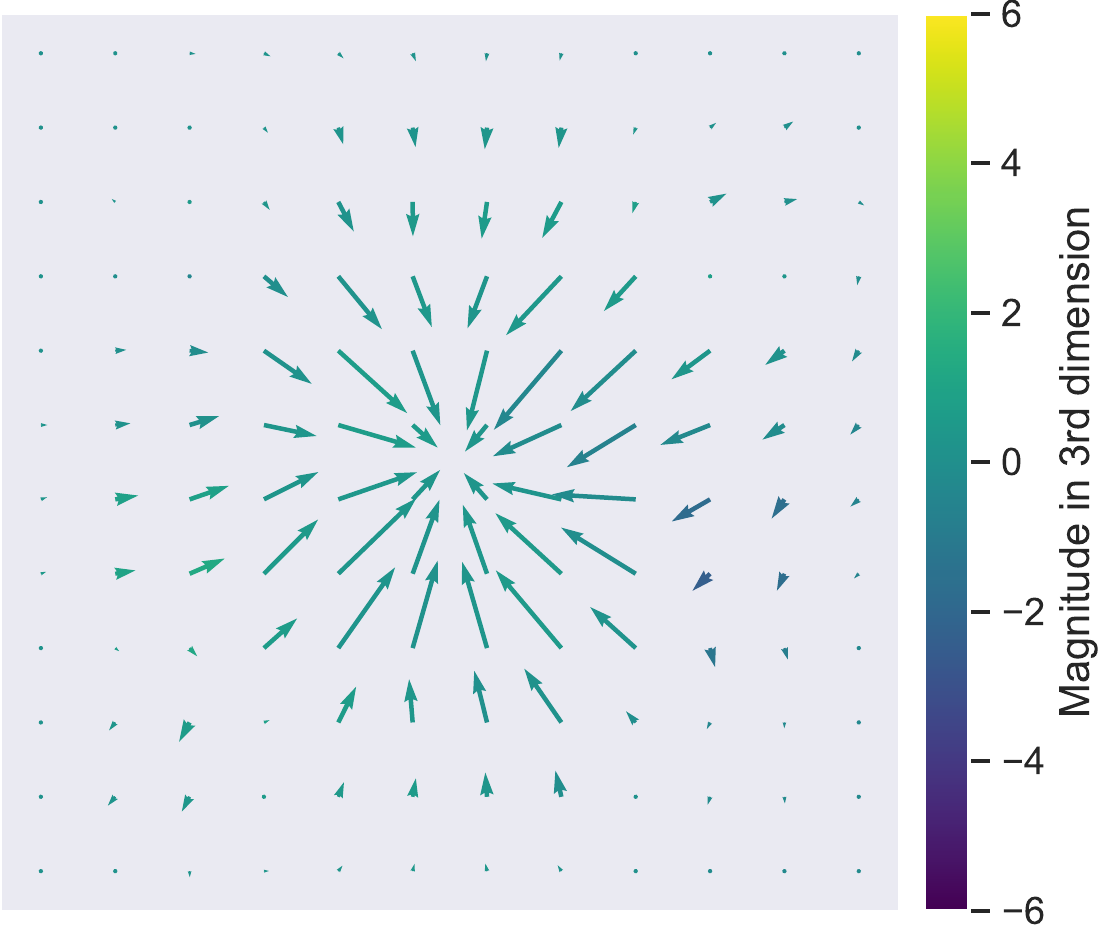}}\hspace{1cm}\quad
    \subfloat[\centering Solution:\newline Approximation front]{\includegraphics[height=\dvfHeight,valign=c]{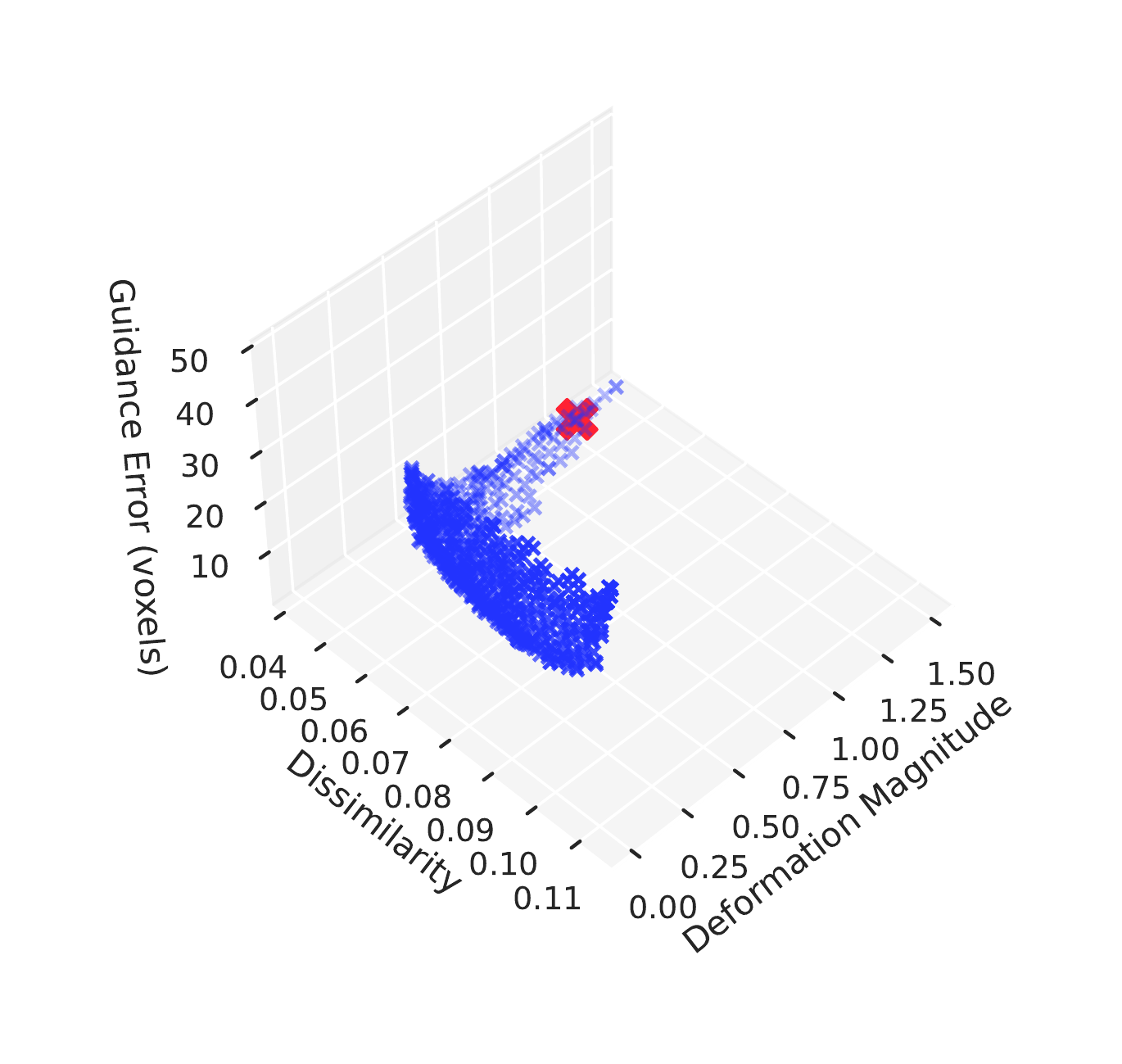}}%
    \caption{The solution with the lowest guidance error in the approximation set produced by the proposed registration method on a synthetic registration problem. A cross-section is taken at $y=25$ for inspection into the volume. The transparency of each voxel represents its grey-value. The Deformation Vector Field (DVF) slice is taken at $z=25$. The selected solution is highlighted in red in the approximation front.}%.\vspace{-0.55cm}}
    \label{fig:images-synthetic}
\end{figure}

% \newcommand{\imageHeight}{3.22cm}
% \newcommand{\dvfHeight}{3.22cm}
% \begin{figure}
%     \centering
%     \subfloat[\centering Problem:\newline Source image]{\includegraphics[height=\imageHeight,valign=c]{figures/synthetic/source_cropped.pdf}%
%     \vphantom{\includegraphics[height=\dvfHeight,valign=c]{figures/synthetic/dvf.pdf}}}\quad
%     \subfloat[\centering Problem:\newline Target image]{\includegraphics[height=\imageHeight,valign=c]{figures/synthetic/target_cropped.pdf}%
%     \vphantom{\includegraphics[height=\dvfHeight,valign=c]{figures/synthetic/dvf.pdf}}}\quad
%     \subfloat[\centering Solution:\newline Transformed source]{\includegraphics[height=\imageHeight,valign=c]{figures/synthetic/transformed_cropped.pdf}%
%     \vphantom{\includegraphics[height=\dvfHeight,valign=c]{figures/synthetic/dvf.pdf}}}\quad
%     \subfloat[\centering Solution:\newline Axial DVF Slice]{\includegraphics[height=\dvfHeight,valign=c]{figures/synthetic/dvf.pdf}}%
%     \caption{The solution with the lowest guidance error in the approximation set produced by the proposed registration method on a synthetic registration problem. A cross-section is taken at $y=25$ for inspection into the volume. The transparency of each voxel represents its grey-value. The Deformation Vector Field (DVF) slice is taken at $z=25$.}%.\vspace{-0.55cm}}
%     \label{fig:images-synthetic}
% \end{figure}

\begin{figure}[t]
    \centering
    \subfloat[\centering Problem:\newline Source -- full bladder]{\includegraphics[height=\imageHeight,valign=c]{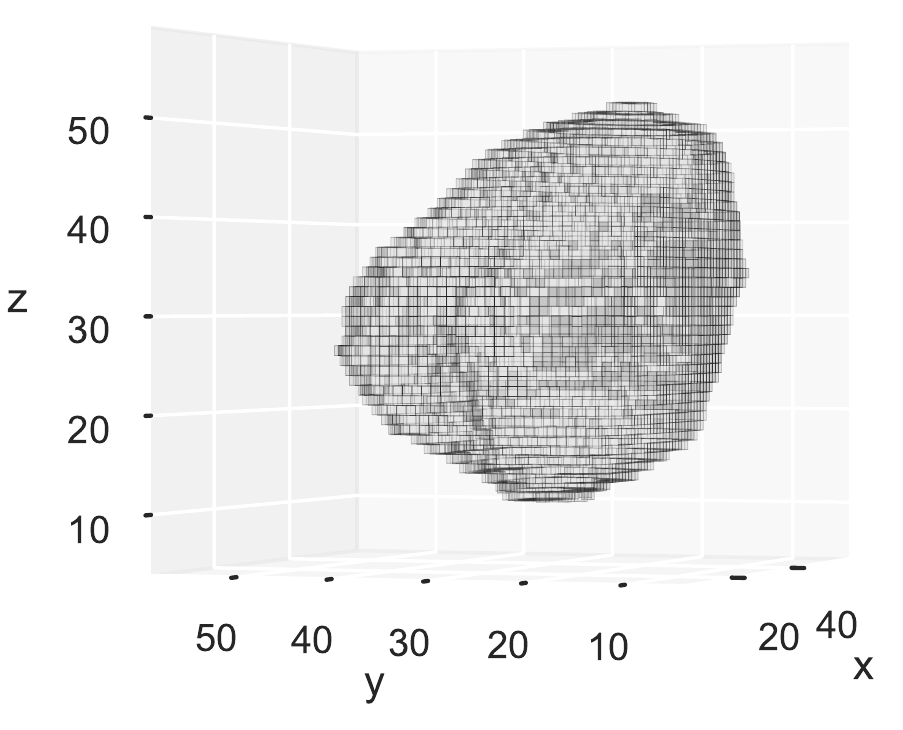}}\quad
    \subfloat[\centering Problem:\newline Target -- empty bladder]{\includegraphics[height=\imageHeight,valign=c]{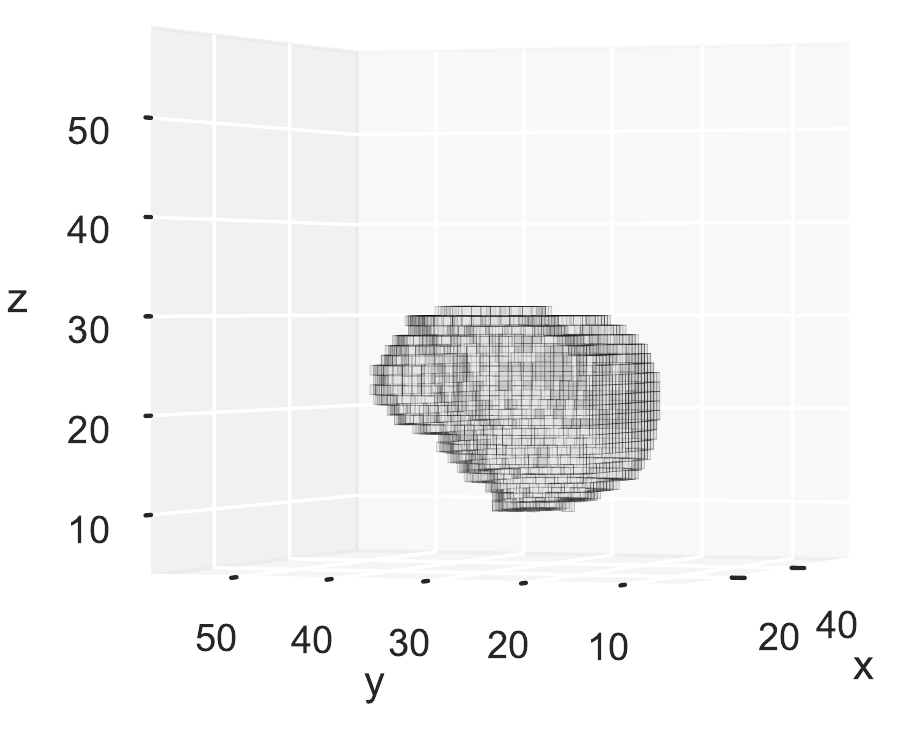}}\quad
    \subfloat[\centering Solution:\newline Transformed source]{\includegraphics[height=\imageHeight,valign=c]{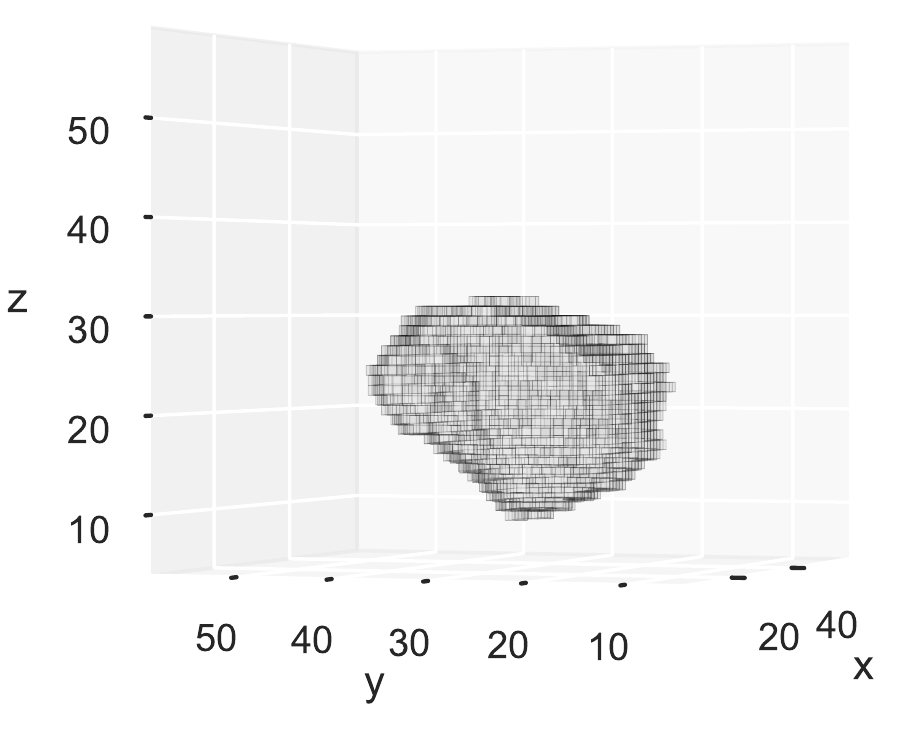}}\quad
    \subfloat[\label{fig:images-real:dvf}\centering Solution:\newline Axial DVF Slice]{\includegraphics[height=\dvfHeight,valign=c]{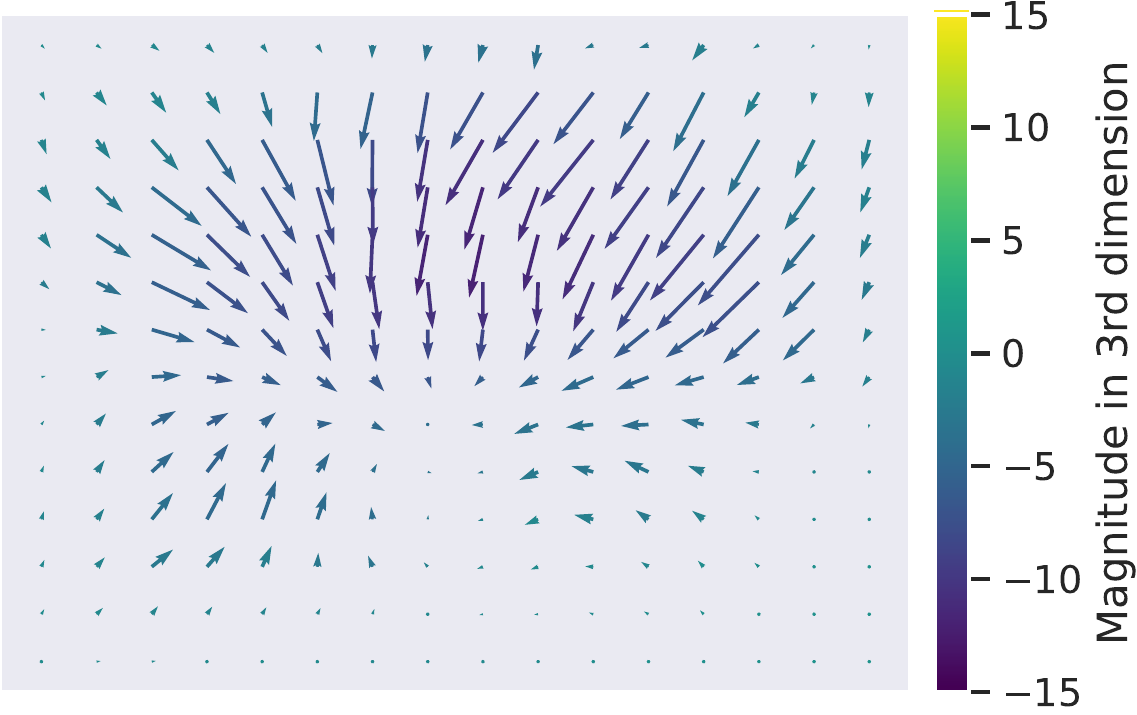}}\hspace{1cm}\quad
    \subfloat[\label{fig:images-real:approx-front}\centering Solution:\newline Approximation front]{\includegraphics[height=\dvfHeight,valign=c]{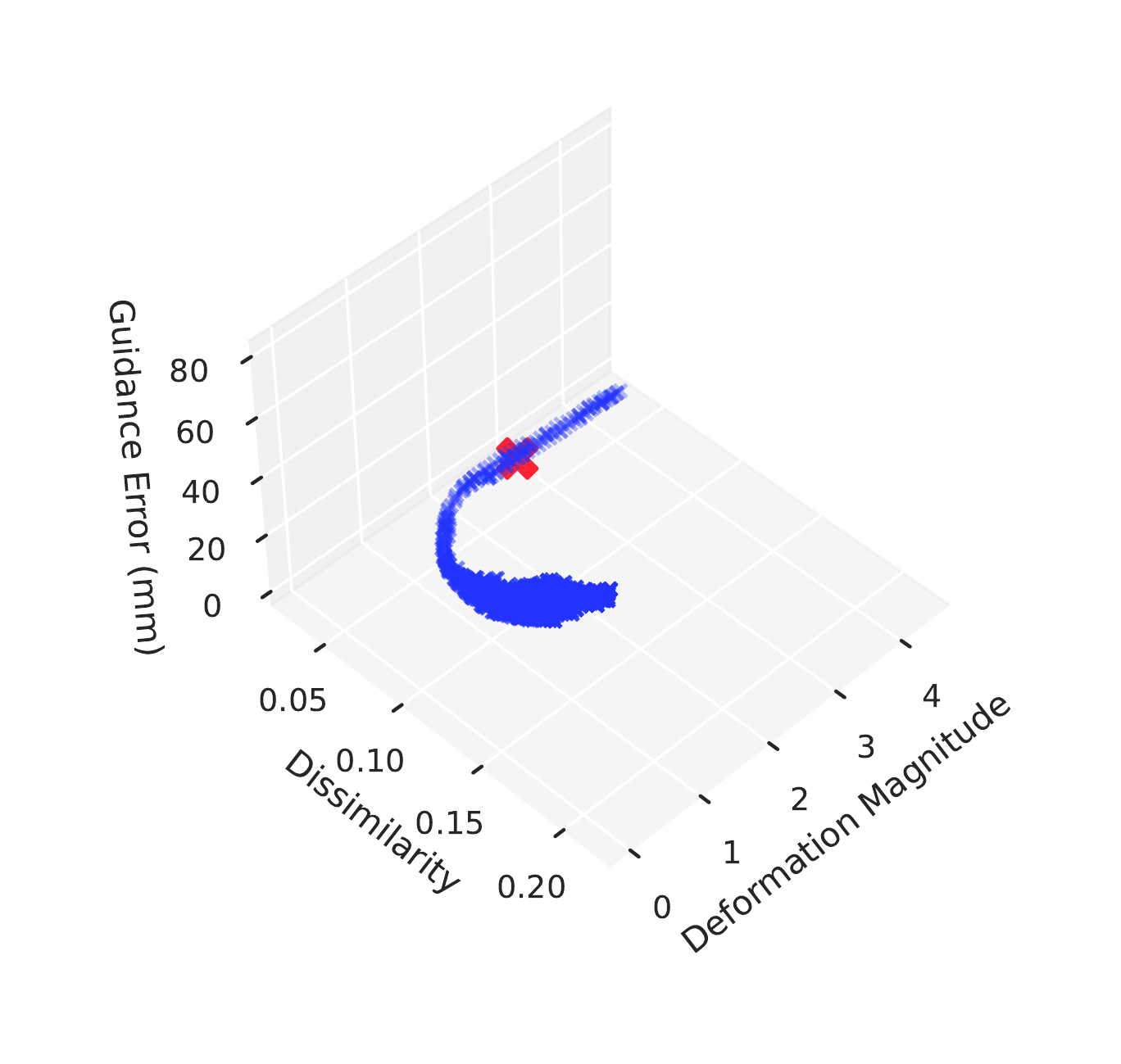}}%
    \caption{The selected solution from the approximation set produced by the proposed registration method on a clinical registration problem. Full volumes are rendered, with the transparency of each voxel representing its grey-value. The Deformation Vector Field (DVF) slice is taken at $z=32$. The selected solution is highlighted in red in the approximation front.}%\vspace{-0.4cm}}
    \label{fig:images-real}
    % \vspace{-0.28cm}
\end{figure}

% \begin{figure}[t]
%     \centering
%     \subfloat[\centering Problem:\newline Source -- full bladder]{\includegraphics[height=\imageHeight,valign=c]{figures/bladder/source_cropped.pdf}%
%     \vphantom{\includegraphics[height=\dvfHeight,valign=c]{figures/bladder/approx-set_cropped.pdf}}}\quad
%     \subfloat[\centering Problem:\newline Target -- empty bladder]{\includegraphics[height=\imageHeight,valign=c]{figures/bladder/target_cropped.pdf}%
%     \vphantom{\includegraphics[height=\dvfHeight,valign=c]{figures/bladder/approx-set_cropped.pdf}}}\quad
%     \subfloat[\centering Solution:\newline Transformed source]{\includegraphics[height=\imageHeight,valign=c]{figures/bladder/transformed_cropped.pdf}%
%     \vphantom{\includegraphics[height=\dvfHeight,valign=c]{figures/bladder/approx-set_cropped.pdf}}}\quad
%     \subfloat[\label{fig:images-real:approx-front}\centering Solution:\newline Approximation front]{\includegraphics[height=\dvfHeight,valign=c]{figures/bladder/approx-set_cropped.pdf}}%
%     \caption{The selected solution from the approximation set produced by the proposed registration method on a clinical registration problem. Full volumes are rendered, with the transparency of each voxel representing its grey-value. The selected solution is highlighted in red in the approximation front.}%\vspace{-0.4cm}}
%     \label{fig:images-real}
%     % \vspace{-0.28cm}
% \end{figure}

\section{RESULTS}
\label{sec:results}

We observe in Figure~\ref{fig:images-synthetic} that our approach can tackle the synthetic registration problem (to within the limits of the used grid resolutions) and can find a desirable transformation for the large deformations present.
Without the use of a multi-resolution scheme this transformation is not found, which underlines the importance of an initial coarse alignment to the success of the overall optimization.
The Deformation Vector Field (DVF) shows that the visual deformation is backed by a transformation that is inverse-consistent, as guaranteed by the prohibition of mesh folding, and visually smooth, as encouraged by the elastic spring model based on Hooke's law.
For the clinical registration problem, Figure~\ref{fig:images-real} illustrates one of the suggested solutions of our registration method, which we selected by taking the solution with the lowest guidance error and navigating towards solutions with lower deformation magnitude up until the guidance error worsens considerably.
This selected solution has a guidance error of $1.149mm$ (slightly higher than the lowest guidance error achieved in the set, $1.099mm$), and a Dice score of 0.911.
The deformation of the bladder is modeled in a smooth fashion and also accounted for in all three dimensions (as inspected from the DVF).

% TOCONSIDER: Smart grid initialization

% \begin{figure}
% \begin{floatrow}
% \ffigbox{%
%   \includegraphics[width=0.45\textwidth]{figures/approx-front.png}%
% }{%
%   \caption{Approximation front produced by one run of the algorithm on patient 1. Mean TRE of each solution in $mm$ indicated by color.}%
%   \label{fig:approx-front}%
% }
% \capbtabbox{%
%     \begin{tabular}{rrr}
%         \toprule
%         Patient & Solution TRE & Baseline TRE \\
%         \midrule

%         1 & 4.58 (2.04) & 5.77 (1.97) \\
%         2 & 10.28 (4.53) & 16.21 (7.02) \\
%         3 & 9.25 (5.19) & 10.52 (8.71) \\
%         4 & 10.33 (3.95) & 17.05 (5.93) \\
%         5 & 8.72 (2.29) & 9.96 (2.92) \\

%         \bottomrule
%     \end{tabular}
%     % \begin{tabular}{rrr}
%     %     \toprule
%     %     Patient & Solution TRE & Baseline TRE \\
%     %     \midrule

%     %     1 & 4.24 (1.44) & 5.78 (2.06) \\
%     %     2 & 8.87 (6.22) & 16.09 (6.92) \\
%     %     3 & 7.08 (5.00) & 9.41 (9.11) \\
%     %     4 & 5.17 (3.65) & 8.40 (2.73) \\

%     %     \bottomrule
%     % \end{tabular}
% }{%
%   \caption{Best mean TREs in $mm$ per patient with standard deviation, and compared against a baseline of only affine registration.}%
%   \label{tab:tres}%
% }
% \end{floatrow}
% \end{figure}

% \vspace{-0.2cm}
\section{DISCUSSION AND CONCLUSIONS}
\label{sec:discussion-and-conclusions}

In this work, we have successfully designed and evaluated a novel multi-objective approach to 3D deformable image registration.
Using a novel 3D dual dynamic grid transformation model, we successfully transferred the existing 2D multi-objective registration approach that is capable to account for large anatomical differences to the 3D image domain.
We have demonstrated the feasibility of this approach through proof-of-concept experiments and foresee promising avenues towards tackling full-scale medical image registration problems.

We see several directions for improvement in future work to make this method capable of solving full-scale clinical registration problems.
First, we see potential in incorporating bio-mechanical modeling, to enhance the likelihood of physically plausible deformations.
This line of work on multi-objective registration already brings together contour-based registration and image-based registration into one extensible optimization model.
As our model shares similarities with typical finite element modeling approaches, bio-mechanical modeling can be further enhanced in a straightforward fashion by assigning material properties to the grid.
Second, changing the image dissimilarity objective to an entropy-based measure could improve resilience to noise in the images.
Third, the quality of registration solutions could be improved by a higher grid resolution.
Currently available GPUs do not have a large enough shared block memory to hold solutions at larger resolutions, however, which calls for more advanced handling and evaluation of registration solutions.
Finally, we are currently conducting larger-scale experiments with full-scale medical scans.
This will also allow for comparisons with other existing methods, using a registration error over expert-placed corresponding landmarks.

\acknowledgments % equivalent to \section*{ACKNOWLEDGMENTS}       

The authors thank E.C. Harderwijk (Dept. of Radiation Oncology, Leiden University Medical Center, Leiden, The Netherlands) for his contributions to this study.
The research is part of the research programme Open Technology Programme with project number 15586, which is financed by the Dutch Research Council (NWO), Elekta, and Xomnia. Further, the work is co-funded by the public-private partnership allowance for top consortia for knowledge and innovation (TKIs) from the Ministry of Economic Affairs.

% References
\bibliography{references} % bibliography data in report.bib
\bibliographystyle{spiebib} % makes bibtex use spiebib.bst

\end{document}